\documentclass{article}
\usepackage[a4paper, margin=1in]{geometry}

\usepackage{hyperref}
\usepackage{graphicx}     
\usepackage{tabularx}
\usepackage{subcaption}
\usepackage{float}
\usepackage{hyperref}
\usepackage{titlesec}
\usepackage{amsfonts}
\usepackage{amsmath}
\usepackage{amssymb}
\usepackage{listings}
\usepackage{xcolor}

\title{\LARGE \bf
	Development of a PPO-Reinforcement Learned Walking Tripedal Soft-Legged Robot using SOFA}

	\author{
		\begin{tabular}{c}
			Yomna Mokhtar$^{1}$, Tarek Shohdy$^{2}$, Abdallah A. Hassan$^{*}$, Mostafa Eshra, \\ Omar Elmenawy, Osama Khalil, Haitham El-Hussieny
		\end{tabular}
	}
	
\begin{document}

		\maketitle
		\thispagestyle{empty}
		\pagestyle{empty}

		\begin{abstract}
			
			Rigid robots were extensively researched, whereas soft robotics remains an underexplored field. Utilizing soft-legged robots in performing tasks as a replacement for human beings is an important stride to take, especially under harsh and hazardous conditions over rough terrain environments. For the demand to teach any robot how to behave in different scenarios, a real-time physical and visual simulation is essential. When it comes to soft robots specifically, a simulation framework is still an arduous problem that needs to be disclosed. Using the simulation open framework architecture (SOFA) is an advantageous step. However, neither SOFA's manual nor prior public SOFA projects show its maximum capabilities the users can reach. So, we resolved this by establishing customized settings and handling the framework components appropriately. Settling on perfect, fine-tuned SOFA parameters has stimulated our motivation towards implementing the state-of-the-art (SOTA) reinforcement learning (RL) method of proximal policy optimization (PPO). The final representation is a well-defined, ready-to-deploy walking, tripedal, soft-legged robot based on PPO-RL in a SOFA environment. Robot navigation performance is a key metric to be considered for measuring the success resolution. Although in the simulated soft robots case, an 82\% success rate in reaching a single goal is a groundbreaking output, we pushed the boundaries to further steps by evaluating the progress under assigning a sequence of goals. While trailing the platform steps, outperforming discovery has been observed with an accumulative squared error deviation of 19 mm. The full code is publicly available at \href{https://github.com/tarekshohdy/PPO_SOFA_Soft_Legged_Robot.git}{github.com/tarekshohdy/PPO$\textunderscore$SOFA$\textunderscore$Soft$\textunderscore$Legged$\textunderscore$ Robot.git}
			
		\end{abstract}

		\section{Introduction}
		The interest in legged robotics has grown significantly due to its ability to traverse diverse terrains and perform complex locomotion tasks. Numerous studies have demonstrated the remarkable capabilities of traditional bipedal and quadrupedal robots in controlled environments. However, soft-actuated legged robots controlled via reinforcement learning (RL) remain underexplored. Tripedal locomotion presents unique challenges, including maintaining dynamic stability, generating asymmetric gaits, and ensuring balance. Additionally, the use of soft actuators introduces further complexity, making this a compelling area of study for researchers in both robotics and artificial intelligence fields.
		
		Soft-legged robots offer a multitude of advantages in both compliance and flexibility, which are particularly well-suited for unstructured environments and uneven terrain traversing applications \cite{miao2017softrobot}. The nonlinear dynamics inherent in soft materials introduce significant control challenges. These uncertainties necessitate data-driven approaches, such as RL. Recent advancements in Proximal Policy Optimization (PPO) and other RL methods have enabled robots to learn stable locomotion policies without explicit mathematical models \cite{kim2024leggedrobot}. Integrating RL with simulation frameworks such as SOFA (Simulation Open Framework Architecture) enables significant advancements by providing realistic environments for policy training prior to real-world deployment \cite{menager2022sofagym}.
		
		In our research, we design and develop a conceptual tripedal robot and train it using PPO within the SOFA framework and address key challenges, including gait generation, balance control, and sim-to-real transfer. By extending current RL-based locomotion architectures, we aim to advance the understanding of soft tripedal locomotion and demonstrate its potential for real-world applications.
		
		This paper is organized as follows: Section \href{II}{\ref{Related Work}} reviews some related work on the general (rigid and soft)-legged robots, legged robot locomotion control strategies, training legged robots using RL, and prior works delving into employing soft-legged robots inside simulation frameworks. Section \href{III}{\ref{Methodology}} presents the proposed methodology, including our main approach key points including the SOFA simulation framework to build the Gym environment to train the model and the utilization of PPO as a SOTA algorithm in RL to generate a walking sequence. Section \href{IV}{\ref{Results}} illustrates the final results collected from the trained robot agents inside SOFA after testing the robot using one simple multiple navigation goals and another complicated assessment to follow a certain path. Finally, Section \href{V}{\ref{Conclusion}} concludes the paper work and outlines several future research directions and capabilities.
		
		\section{Related Work}
		\label{Related Work}
		\subsection{Rigid vs Soft Legged Robots}
		Rigid-legged robots are made of hard, inflexible materials and have characteristics like high precision, payload capacity, and ease-of-control \cite{siciliano2010robotics}. They work excellently in applications where structured tasks with high accuracy and heavy loads are required. At the same time, they do not have much flexibility and adaptability to the dynamic or unstructured environment. On the other hand, stiff actuation means conventional electric motors or hydraulic systems. They provide high power and great precision but may lack compliance or flexibility. 
		
		Meanwhile, soft-legged robots are made from soft, deformable materials and thus are able to adapt to their environment. They have been inspired by biological organisms and leverage their intrinsic compliance and adaptability for the navigation of challenging environments and interactions with humans and objects \cite{kim2013softrobots}. Soft robots move and deform through new ways of actuation like pneumatic or hydraulic methods. Although it has many safety advantages and adaptabilities, one of the major barriers of soft robots is controllability, precision, and stability due to their deformed nature \cite{calisti2012softrobot}  \cite{trivedi2008softrobot}.
		
		\subsection{Legged Locomotion and Control Strategies}
		Legged robots are gaining traction because they can traverse intricate and unstructured terrains better than wheeled or tracked systems. Historically, legged locomotion control strategies have employed Central Pattern Generators (CPG), Model Predictive Control (MPC), and Zero-Moment Point (ZMP)-based techniques.
		
		Borrowing from CPG-based controllers, the organisms whose movement is modeled walk with a rhythmic pattern that is adaptive and smooth thanks to tuned oscillators that generate gait patterns through oscillation. They tend to be very difficulty terrain adaptive \cite{ijspeert2008cpg}. On the other hand, foot placement optimization is easily tackled by MPC while also being applied widely on quadrupedal and bipedal locomotion centers \cite{inoue2024tripedalrobot}. With the advent of non symmetrical and underactuated robotic configurations like tripedal robots, MPC has been deemed expensive and ineffective for computation.
		
		These approaches target dynamic motion by maintaining balance under motions making sure that the resultant force is within a dynamically stable zone of the polygon of support \cite{kajita2003biped}. While successful in walking robots’ ASIMO and Atlas, they find a shortcoming with dynamic and underactuated robotics where it is difficult to ensure ZMP is preset at all times.
		
		Recent studies suggest that reinforcement learning (RL) offers a promising alternative, particularly for low-rigidity configurations where conventional controllers struggle to ensure stability \cite{inoue2024tripedalrobot}.

		\subsection{Reinforcement Learning for Legged Robots}
		Reinforcement learning has revolutionized robotic mobility by enabling robots to learn complex motion strategies without explicit programming. \cite{kim2024leggedrobot} introduced a barrier-based RL reward function to facilitate learning heterogeneous gaits, including tripedal, bipedal, and quadrupedal walking. Similarly, \cite{kaymak2023humanoidrobot} demonstrated that Deep Reinforcement Learning (DRL) can improve humanoid robot stability through optimized walking trajectories using Dueling Double Deep Q Networks (D3QN). These advancements highlight the potential of RL for addressing the challenges of tripedal locomotion.
		
		Reconfigurable robotic behavior in the line of legged robotic works has recently witnessed adding RL applications. \cite{soccer2024reinforcement} featured adaptive control strategies that would greatly help in demonstrating coordination of multiple agents and on-the-fly decision-making in a dynamic environment, showing how RL is done for reconfigurable robotic soccer. As further work, \cite{modular2024multi} presented a modular RL framework to support increased locomotion efficiency of a multi-legged robot in a hierarchical reinforcement learning context.
		
		\cite{policygradient2024robot} additionally presented a control based on RL for robots with strange morphologies. He showed that it works very efficiently with policy gradient methods when optimally formulating movement strategies, even for legged robots whose configuration is non-conventional. \cite{realtime2024robot} thus employed Proximal Policy Optimization at the background of real-time gait adaptation to quadrupedal locomotion such that dynamic stability and variation due to the environment are catered for in the constraints used. 
		
		Otherwise, \cite{safereinforcement2024softrobot} are more concerned with safe RL approaches for soft robotic locomotion and use constrained policy optimization; they allow seriously stable motion and reduce unwanted deformation in soft-legged robots. In the same vein, \cite{curriculum2024ppo} had curriculum learning combined with PPO to enable soft robots to improve locomotion whenever called without having to adopt any real-world condition.
		
		\subsection{Soft Robotics and Simulation Frameworks}
		Soft robotics introduces additional challenges, such as material deformation modelling and actuation latency. \cite{miao2017softrobot} proposed matrix displacement-based modelling for soft actuators and demonstrated its application in a three-legged soft robot. However, the lack of accurate simulation tools has hindered progress in this field. To address this, SOFA-based platforms like SofaGym have been developed, providing finite element-based simulations tailored for RL applications in soft robotics \cite{menager2022sofagym}. So far, these papers \cite{adagolodjo2019robust} and \cite{ataka2023inverse} utilized the Finite-Element Methods (FEM) for modeling examples of soft robots using the SOFA framework. Both papers use Artificial Neural Network (ANN) to create controllers for the soft robots in order to navigate to specific coordinates. However, it is time-consuming to collect enough data to train the ANN. The RL approach in soft robot control is efficient in taking steps forward to directly create high-level robust controllers, as recommended in the future work of both papers recently mentioned.
		
		However, after many attempts of searching inside the community contributions, there is no prior work providing a navigation of an autonomous mobile soft-legged robot inside SOFA simulation framework already trained on the SOTA of RL, which is the PPO technique. Here our contribution comes, especially because of our utilization of the minimum needed number of legs of the soft robot. Not working with a bipedal to avoid delving into a self-balancing dilemma, nor quadruped or more to get rid of learning over-calculation and data acquisition.
		
		\section{Methodology}
		\label{Methodology}
		\subsection{Real-Time Physical Environment Using SOFA}
		In this section, we will explore the use of the Simulation Open Framework Architecture (SOFA) \cite{faure2012sofa}, an open-source framework primarily used for real-time medical simulation. After reviewing several simulation tools, SOFA has qualified as the best option to provide a realistic simulation environment \cite{coevoet2017softrobotics} for a tripedal soft-legged robot to train to stabilize itself while walking to reach a sequence of given goals, which hasn't been implemented before on SOFA. We wanted to uniquely adapt SOFA's full potential to model our tripedal soft-legged robot by utilizing its advanced dynamic capabilities and complex physical and behavioral modeling options. The first distinct aspect of SOFA that sets it apart from any other simulation tool is that it provides a wide variety to model different materials by defining the simulated object's internal forcefields through a set of parameters, which is ideal for our project to enable us to closely model the same Thermoplastic Polyurethane (TPU) material that was used to fabricate the tripedal robot's legs. 
		
		Another important aspect of SOFA is external forcefields, which define forces in the environment, such as gravity and weight. These forcefields showcase the interaction between different objects in the defined environment, opening the door to exploring more applications of SOFA for autonomous robots that navigate different terrains and dynamically interact with them.
		
		Accordingly, we are going to review the process of creating the robot's leg scene in SOFA with cable actuation inputs and end-effector position feedback using \textit{SofaPython3} plugin for Python scripts to essentially provide the capability to integrate the created scene later with other Python libraries for Reinforcement Learning (RL). Then, a scene with the full robot's model will be created in the same manner.
		
		\subsubsection{Leg Scene} \label{sec:leg_scene}
		We started customizing our scene in SOFA, which should first model the soft leg actuated with cables. To do so, we need to specify several parameters, starting with the material selection, how to fix the leg, what to use for actuation, and how to receive feedback on its position.
		
		First, we set the default \lstinline[language=Python]{CGLinearSolver}
		and \lstinline[language=Python]{EulerImplicitODESolver}
		and defined the scene's time step and external force field which is the gravity and the mass for the weight. Then, we started with the material selection process, which was tedious after going through numerous trials to find the best material that closely simulated our physical soft leg's behaviour. We found the cable-driven gripper project implemented using \textit{SoftRobots} plugin \cite{coevoet2017softrobotics} and used a custom object created from the \textit{STLIB} plugin that is called \lstinline[language=Python]{ElasticMaterialObject}. It uses Young's modulus and Poisson ratio parameters along with the \textit{VTK} volume mesh file and \textit{STL} surface mesh file to define the material behaviour and provides much more satisfactory results than that with the other trials. As a consequence, we implemented it as a part of our leg scene to define it and used \textit{STLIB} custom mapping for collision and visuals. The Young's modulus and Poisson ratio parameters were optimized for a more accurate modelling approach that closely matches the physical leg's dynamic behaviour using system identification methods.
		
\begin{figure}[!b]
			\centering
			\includegraphics[width=0.45\textwidth]{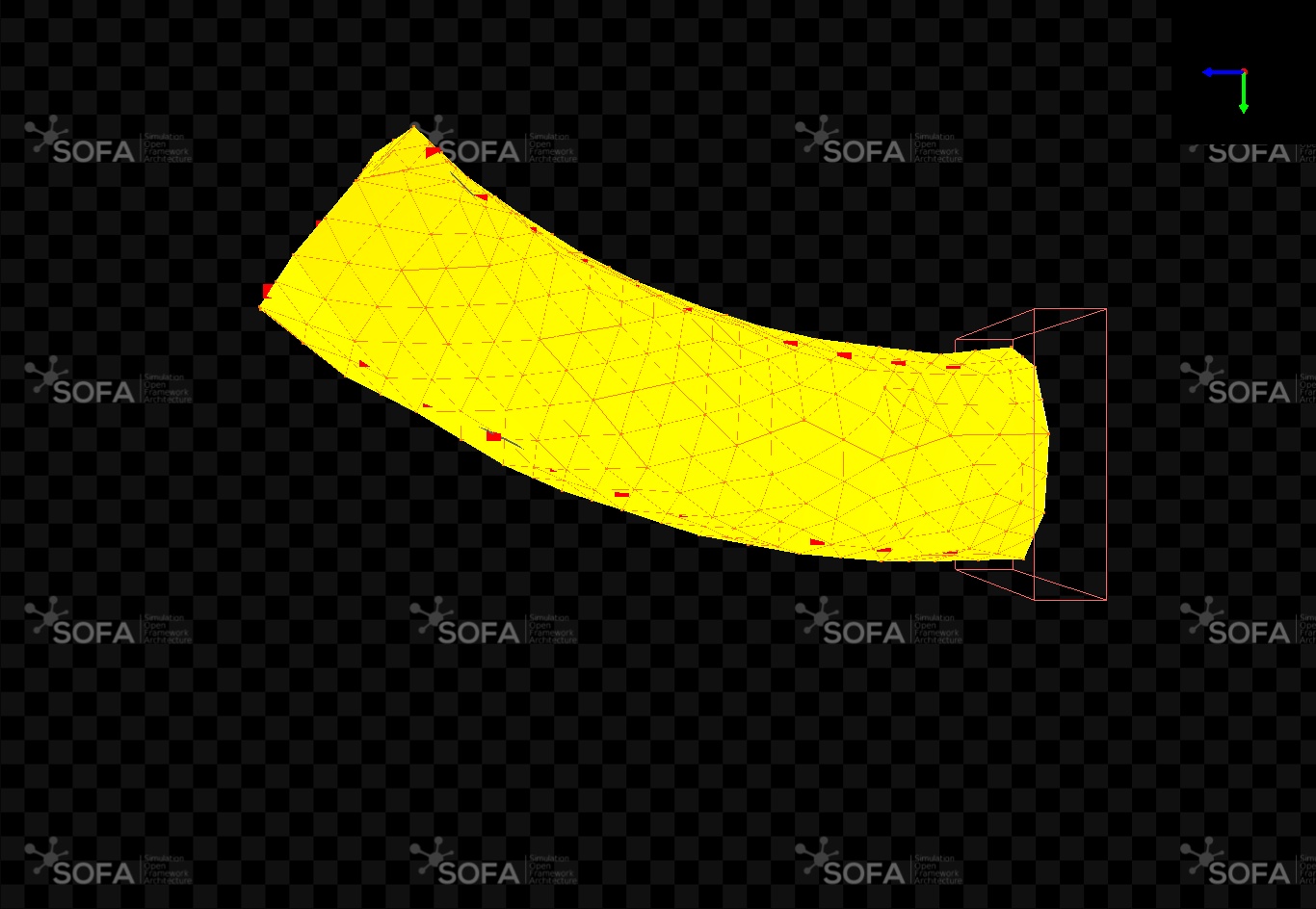}
			\caption{The leg bending as the cables displacements change}
			\label{fig:legbentsofa}
		\end{figure}

		For the actuation, we used the cable actuators function \lstinline[language=Python]{PullingCable()} defined in the \textit{SoftRobots} plugin and created in \cite{adagolodjo2021softrobotics} that required a \textit{JSON} file defining the cable's geometry. Then, a controller function is used to access the value attribute of the cable object, which defines its length and changes it, deforming the leg according to the input displacements as desired, as in fig. \ref{fig:legbentsofa}.
		
		Three cables were placed uniformly across the leg's cross-section and the lengths of the cables were controlled using \lstinline[language=Python]{LegController()} function. Then, to fix the leg, we used a  
		\lstinline[language=Python]{FixedBox}
		constraint from the \textit{STLIB} plugin with defined indices to keep the leg's tip fixed to the origin against gravity as it moves. 
		
\begin{figure}[!t]
			\centering
			\begin{subfigure}[h]{0.2\textwidth}
				\centering
				\includegraphics[width=\linewidth]{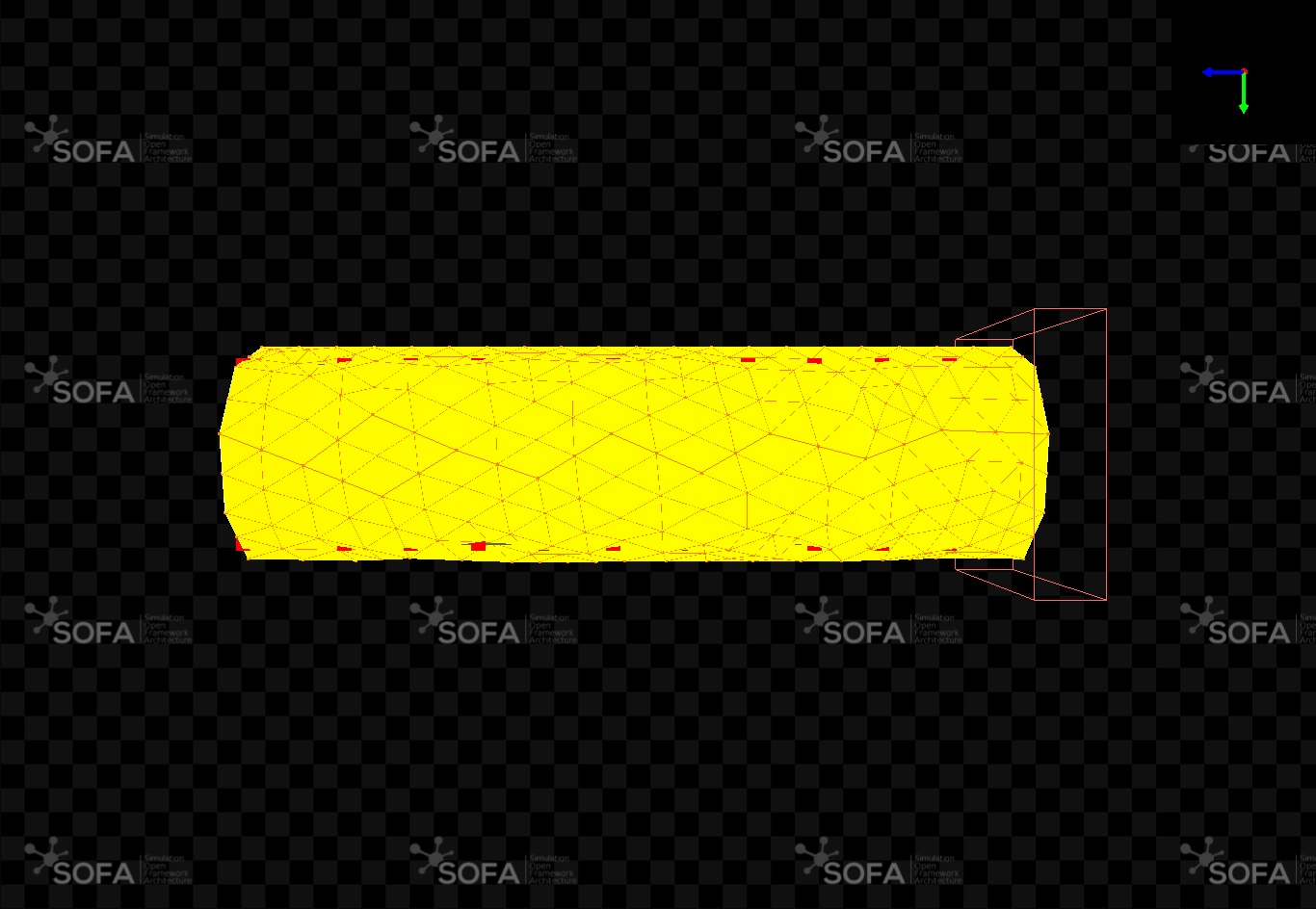}
				\caption{Collision view}
				\label{fig:leg_collision}
			\end{subfigure}
			\hspace{1cm} 
			\begin{subfigure}[h]{0.2\textwidth}
				\centering
				\includegraphics[width=\linewidth]{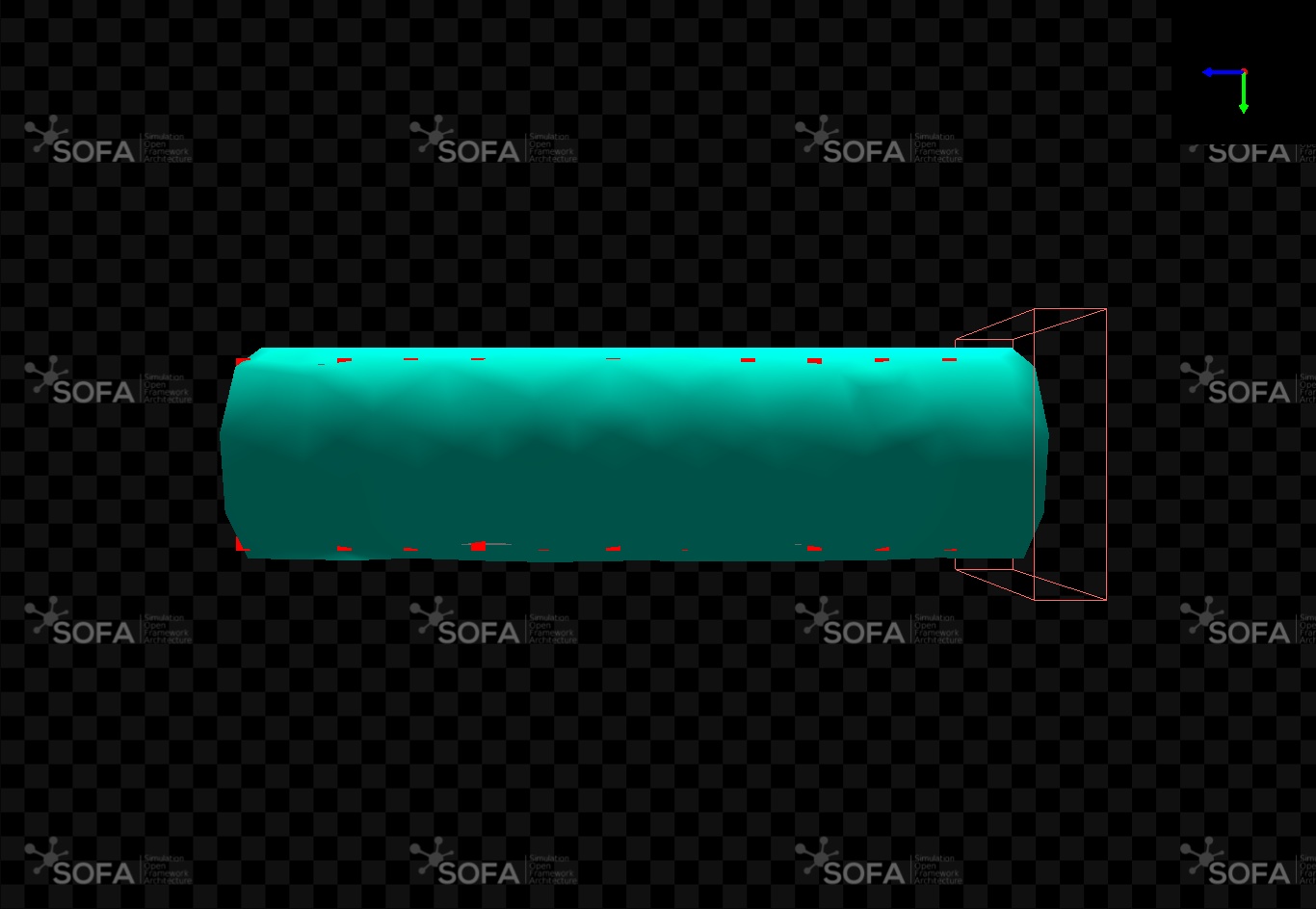}
				\caption{Visual view}
				\label{fig:leg_visual}
			\end{subfigure}\\ 
			\vspace{0.5cm} 
			\begin{subfigure}[h]{0.2\textwidth}
				\centering
				\includegraphics[width=\linewidth]{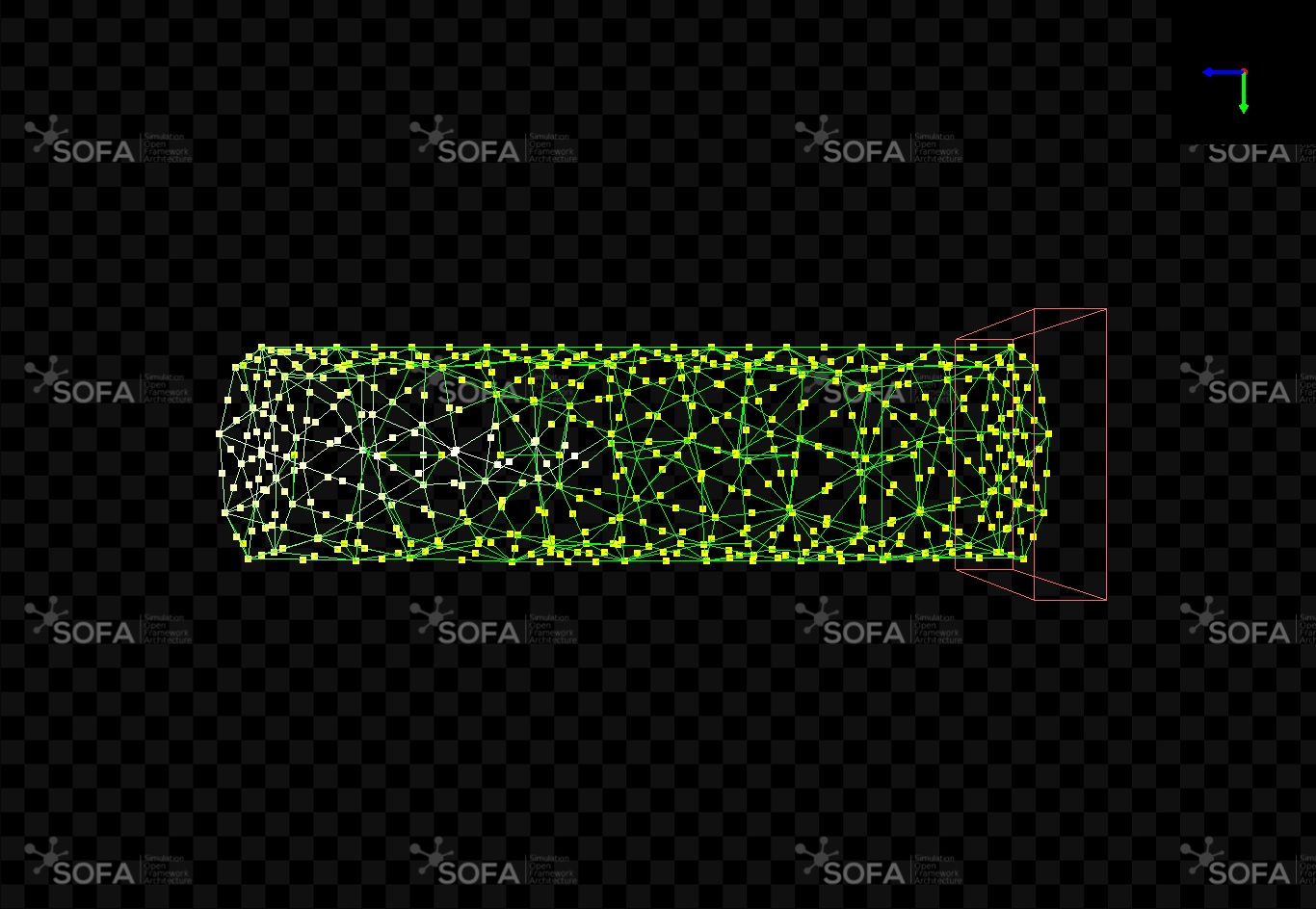}
				\caption{Mappings view}
				\label{fig:leg_mappings}
			\end{subfigure}
			\hspace{1cm} 
			\begin{subfigure}[h]{0.2\textwidth}
				\centering
				\includegraphics[width=\linewidth]{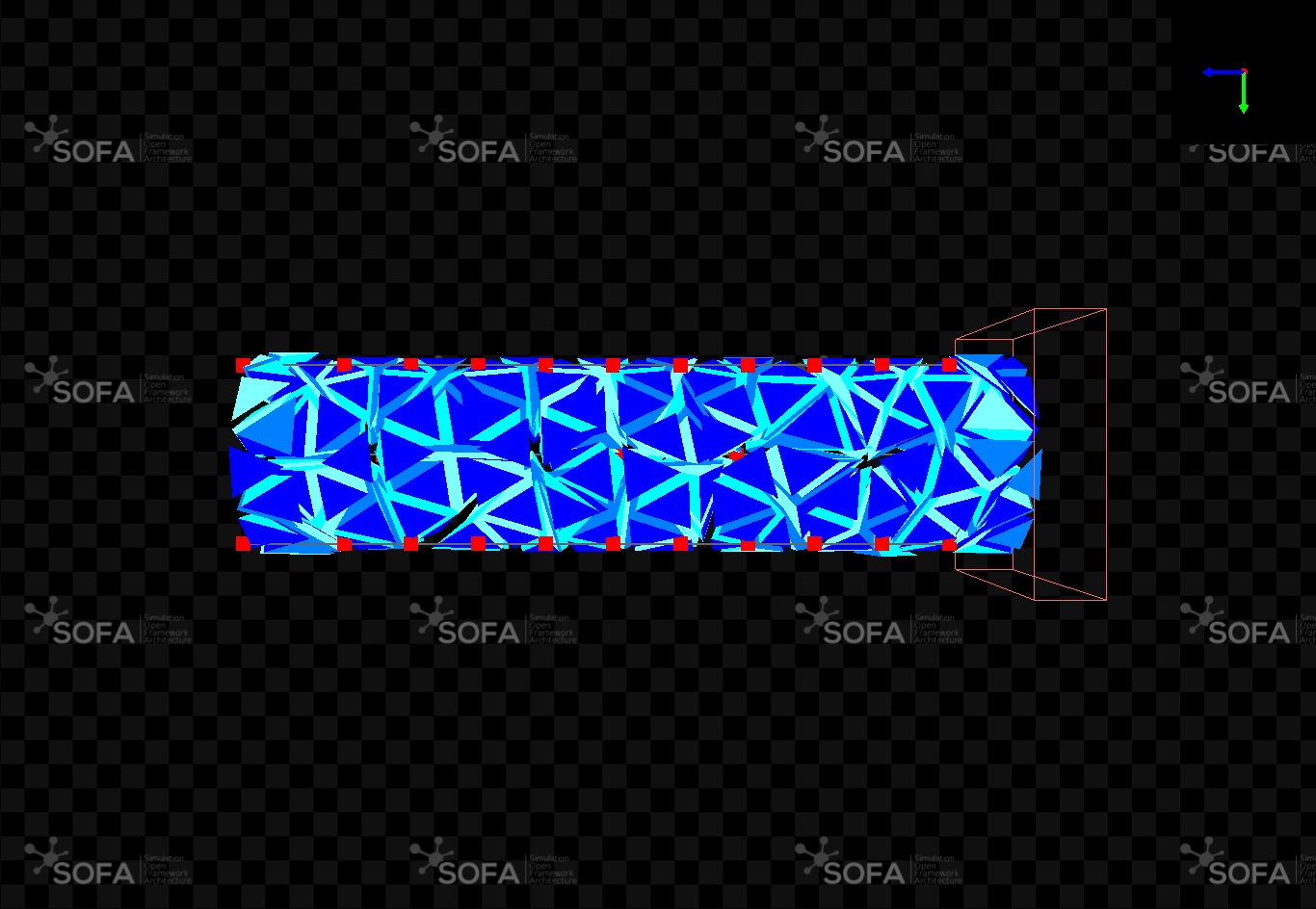}
				\caption{Force field view}
				\label{fig:leg_forcefields}
			\end{subfigure}
			\caption{Leg scene in SOFA}
			\label{fig:legsofa}
		\end{figure}

		Finally, the last step was to get feedback from the leg's end-effector tip with its changing position. Normally, default \lstinline[language=Python]{MechanicalObject}  has position attributes that can be accessed through the Python script as desired. However, since we used custom 
		\lstinline[language=Python]{ElasticMaterialObject} 
		from \textit{STLIB} plugin, it didn't have the same attributes documented which posed a challenge. Consequently, we had to use the collision of the leg which is a \lstinline[language=Python]{MechanicalObject}
		and has the position attribute accessible and since it is mapped to the leg object as it moves, it would accurately provide the needed leg end-effector position. With that, the scene of the leg was finally created, as illustrated in fig. \ref{fig:legsofa}, with all the requirements that satisfy a physical simulation of our model that highly matches its behaviour with precision and accessibility to take actions and keep track of the leg's changing state in the Python script to be deployed later as an environment for reinforcement learning.
		
		\subsubsection{Robot Scene} \label{sec:robot_scene}
\begin{figure}[!t]
			\centering
			\begin{subfigure}[b]{0.22\textwidth}
				\centering
				{\includegraphics[width=\linewidth]{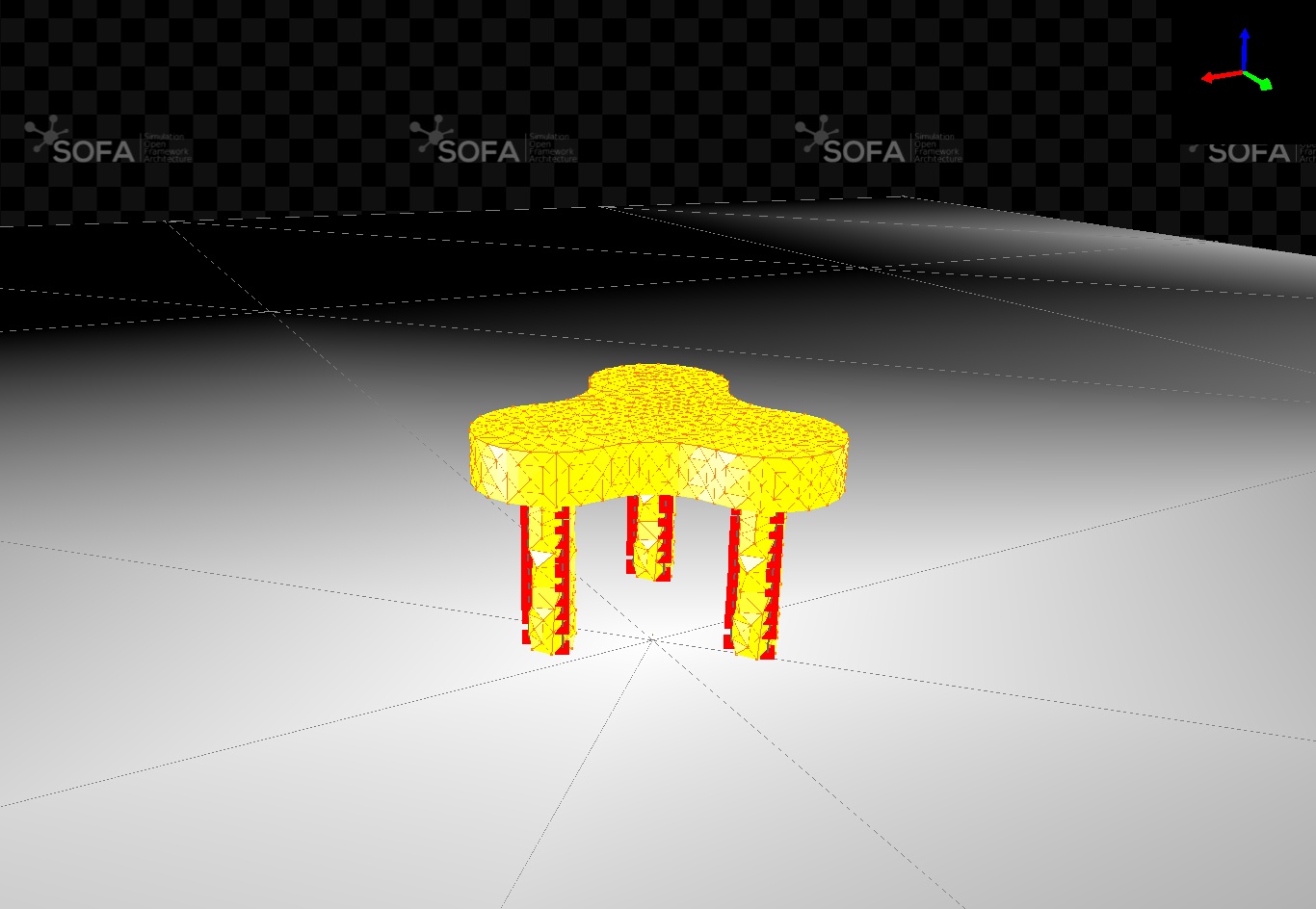}}
				\caption{Collision view}
			\end{subfigure}
			\qquad
			\begin{subfigure}[b]{0.22\textwidth}
				\centering
				{\includegraphics[width=\linewidth]{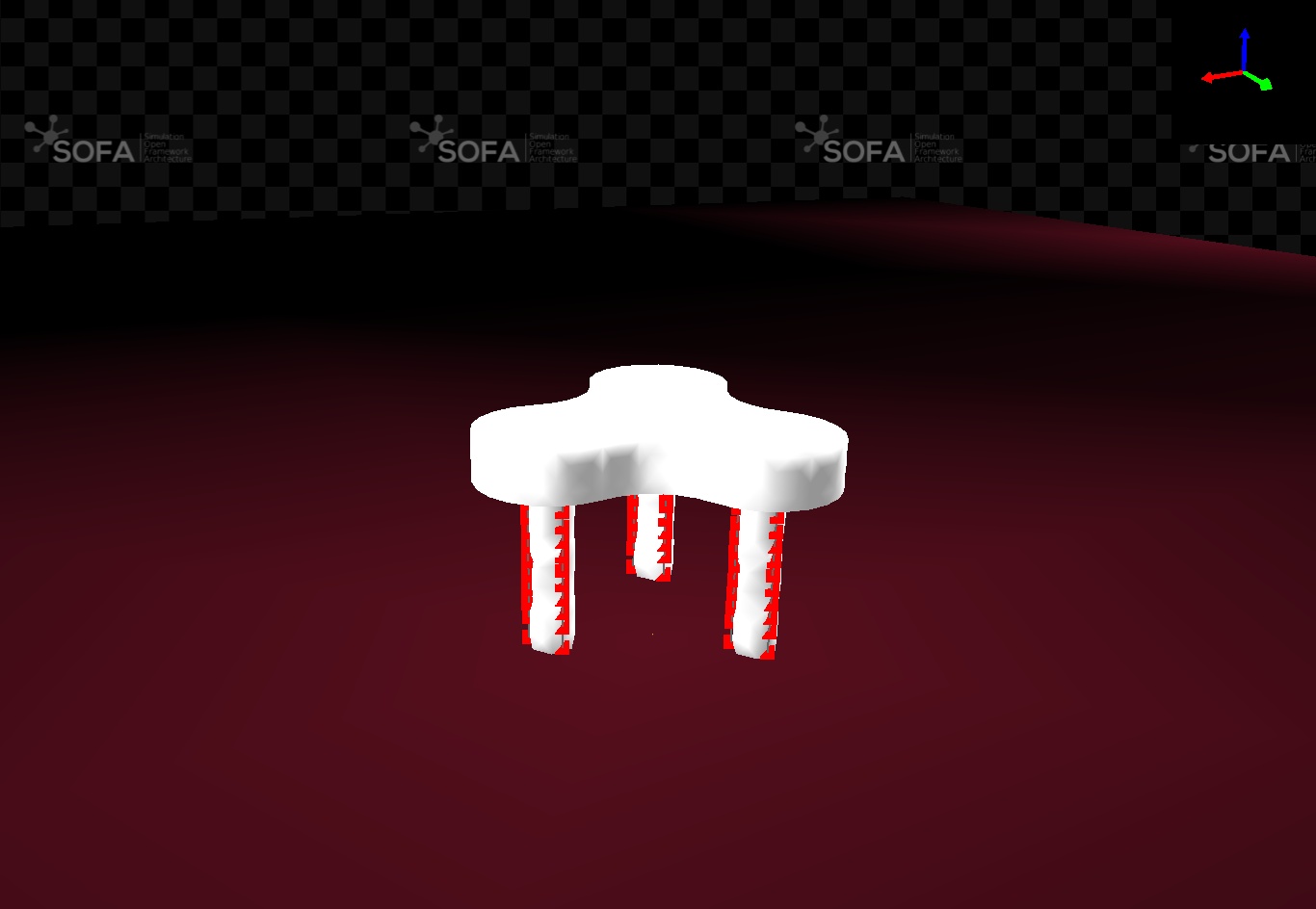}}
				\caption{Visual view}
			\end{subfigure}
			\caption{Robot scene in SOFA}
			\label{fig:robotsofa}
		\end{figure}

		After setting up the leg scene in SOFA, it was much easier to do the same thing, but with the whole robot model where we can actuate the cables of each of the three legs independently and get feedback of the robot's end-effector pose. We started by creating \lstinline[language=Python]{Robot()}
		function to define the robot object similarly to the leg. A simplified design for the robot was created to generate the \textit{STL} and \textit{VTK} mesh files with minimal details to optimize the processing effort it takes to create the SOFA scene. The same collisions and \lstinline[language=Python]{ElasticMaterialObject}
		were used from the \textit{STLIB} plugin to define the robot's material with similar Young's modulus and Poisson ratio as that of the leg. Additional cables were added with their controllers to each of the robot's three legs and were actuated with the same method. However, we added a fixed floor \lstinline[language=Python]{MechanicalObject}
		along with the gravity as an external forccefield to simulate the robot walking on the ground. The additional challenge in this scenario was to optimize the \lstinline[language=Python]{alarmDistance}
		, \lstinline[language=Python]{contactDistance}
		, and \lstinline[language=Python]{frictionCoef}
		parameters in the \lstinline[language=Python]{ContactHeader()}
		function to better simulate how the robot interacts with the ground and walks in real-life. After some trial and error, we reached fulfilling results with these parameters that better suited the real-life robot. Consequently, the robot scene was created as in fig. \ref{fig:robotsofa} and tested by changing the cable values and moving the robot around to verify its behaviour and interactions within the environment.
		\vspace{0.8cm}
		
		\subsection{Reinforcement Learning with Proximal Policy Optimization (PPO-RL)}
		A \textbf{Gym} environment was integrated with the SOFA simulator, scene was created in sec. \ref{sec:robot_scene}. It provides a platform for the reinforcement learning agent to interact with and learn the walking sequence and also, it exhibits chaotic behaviours that challenge the learning process. The environment aims to train the agent policy model to reach a certain goal in the environment depending on the robot's current position.
		
		\begin{enumerate}
			\item \textbf{Observation Space:} A dictionary that includes the 2 keys and values representing:
			\begin{itemize}
				\item \textbf{Agent Location and Rotation:} A continuous space represented by a 6-dimensional vector $(x, y, z, \alpha, \beta, \gamma)$, indicating the current state of the robot whether it's far from the ground or tilted about the $x-axis$, etc... 
				\item \textbf{Target Location:} A 2-dimensional vector $(x_g, y_g)$, representing the desired position the agent aims to reach and is randomized at the beginning of each episode in a specific range $100 > |x_g| > 40, 100 > |y_g| > 40$. 
				
			\end{itemize}
			This setup provides the agent with essential information about its current state and the goal it needs to achieve.
			
			\item \textbf{Action Space:}
			\begin{itemize}
				\item The action space is continuous, allowing for precise control over the nine cables. Each action corresponds to a specific tension applied to the cables, enabling the agent to explore a wide range of configurations for the 3 legs and the platform above.
			\end{itemize}
			
			\item \textbf{Interaction with SOFA Robot Scene} (shown in fig. \ref{fig:sofa_robot_scene_env}): \textbf{SOFA} Simulator works as a headless interface with the \textbf{Gym} environment. With each \lstinline[language=Python]{step}
			function, the action provides the simulation with the cable inputs $(l_1, l_2, l_3, ..., l_9)$, then the \textbf{SOFA} simulation return with the agent position and rotation. 
			
			\begin{figure}[!htbp]
				\centering
				\includegraphics[width=0.9\linewidth]{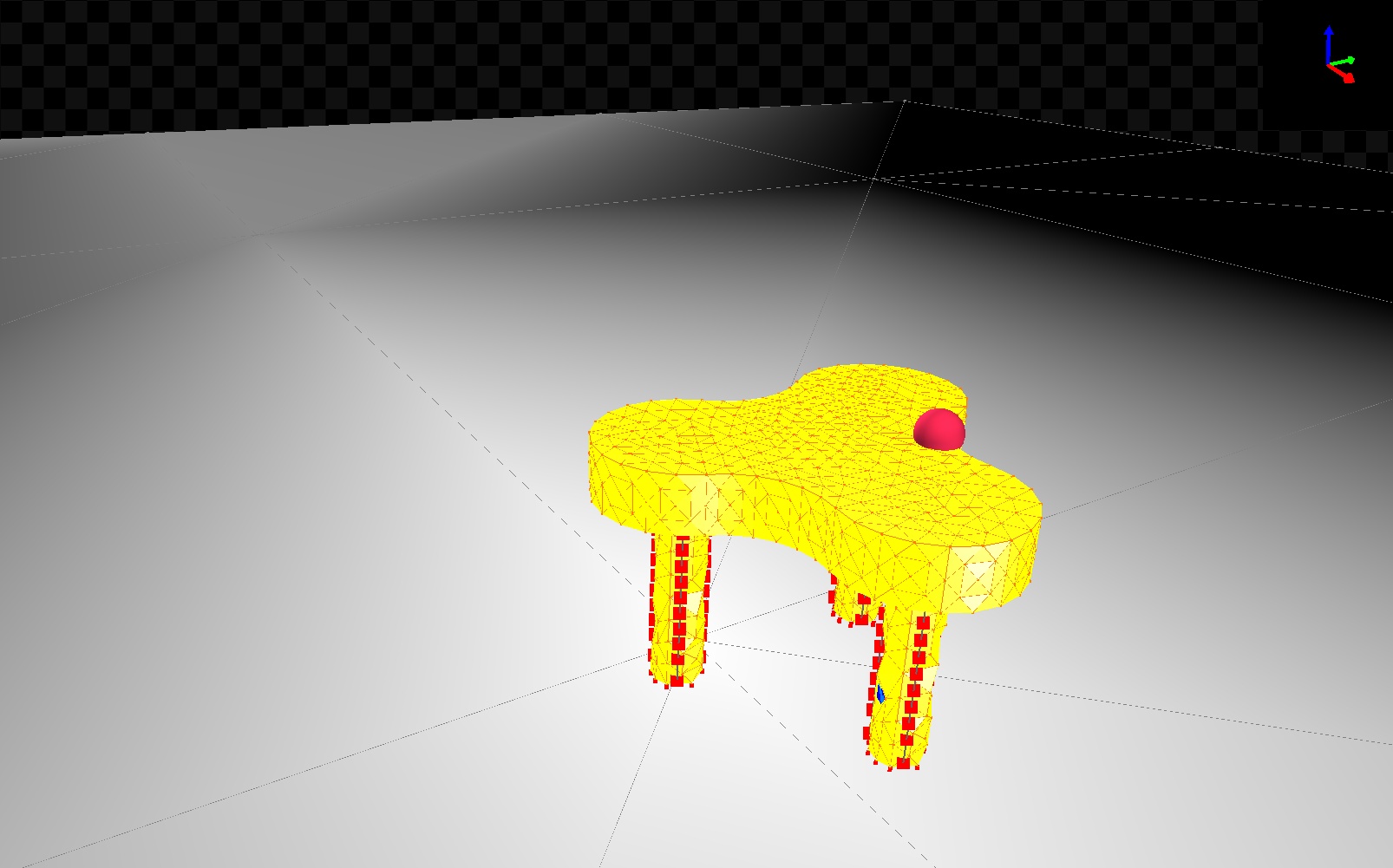}
				\caption{\textbf{SOFA} robot scene environment \textbf{GUI} (The red ball is the goal position visualized with a radius equivalent to the distance threshold)}
				\label{fig:sofa_robot_scene_env}
			\end{figure}
			
			\item \textbf{Termination and Truncation Conditions:}
			\begin{itemize}
				\item \textbf{Termination:} The episode terminates when the agent's location is within a specified threshold distance from the target, indicating successful task completion.
				\item \textbf{Truncation:} The episode is truncated if the agent exceeds a maximum number of time steps (1000), or if the height, roll, or pitch exceeds specified limits, preventing unsafe or failure scenarios.
			\end{itemize}
			These conditions ensure that the training process is both goal-oriented and time-efficient, driving the agent to learn effective strategies for reaching the target location. However, an important aspect for the training to be efficient is the reward function, which won't be straightforward in a complex environment such as this, with the \textbf{PPO} algorithm used, the reward function should be driving the robot to explore more, but with limits to achieve stability and reach goals efficiently. 
			
		\end{enumerate}

		\paragraph{Reward Function Design} \label{sec:reward_fd} The reward function design has gone through a trial and error process to enhance the agent accuracy to reach the desired goal in the environment. The reward used the state of the robot with $d = \sqrt{(x_g-x)^2 + (y_g-y)^2+ (z_g-z)^2}$, $d_{\text{max}}$ is the maximum acceptable error to reach, $\alpha, \beta$ are the roll and pitch of the base of the robot, $v_x, v_y$ are the velocity of the robot in the $x$ and $y$ directions, respectively. The function should reward proximity to the target, penalize excessive roll and pitch, and encourage movement in the correct direction.
		
		\scriptsize
		\begin{equation}\label{eq:reward}
			r = 20 \left( 1 - \frac{\ln(1 + d)}{\ln(1 + d_{\text{max}})} \right) - |\alpha| - |\beta| + v_x \frac{x_g - x}{|x_g - x|} + v_y \frac{y_g - y}{|y_g - y|}
		\end{equation}
		\normalsize
		
		Where the values of $\alpha, \beta$ angles are in degrees $ 0 < \alpha, \beta < 360$ to impose a large penalty on the falling of the robot, this is because angles in degrees have larger numerical values compared to radians. The terms $v_x, v_y$ represent velocities in mm/s which are relatively large values for the model to encourage significant updates in the policy network. The direction terms $\frac{x_g - x}{|x_g - x|}$ and $\frac{y_g - y}{|y_g - y|}$ ensure that the model is rewarded when the velocity aligns with the goal direction relative to the robot’s current position. This function encourages the model to move and change its position by increasing the factor of the distance to the goal and the velocity term.
		
		\section{Results}
		\label{Results}
\begin{figure}[!t]
			\centering
			\includegraphics[width=0.9\linewidth]{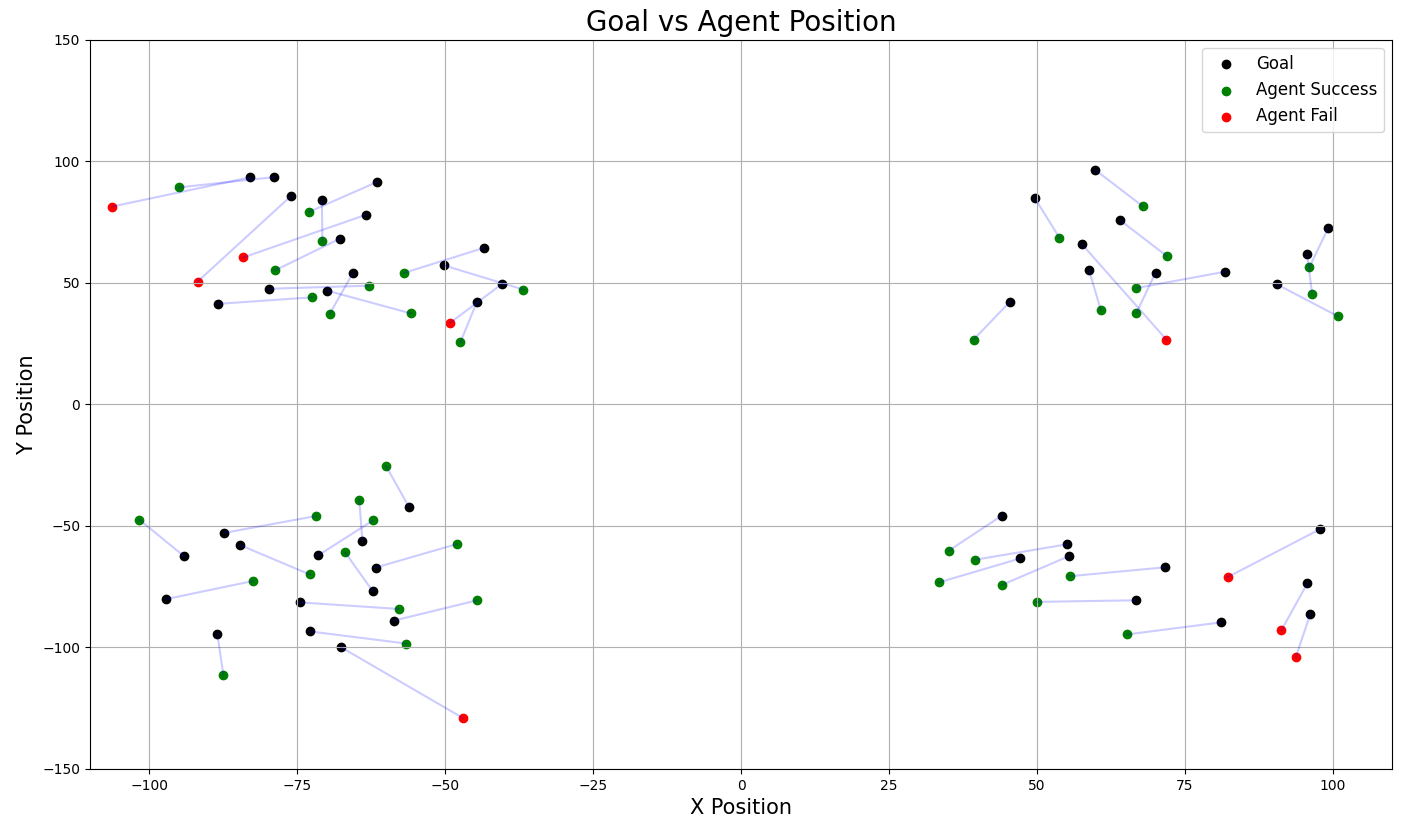}
			\caption{Map of agent best position in each trial and the 50 randomized goals}
			\label{fig:agent_goals}
		\end{figure}

		The test for the best model was executed by giving the agent model 50 random goals to reach in the environment, each one starting from the origin at (0,0) and recording each episode output; did it reach the goal, how many steps it took to reach it and the minimum achieved distance to the goal, the results for this test is shown in tab.\ref{tab:performance_metrics_robot}. Test results came positive with the robot not falling and reaching an accuracy of 82\% with 42 successful times it reached the goal and came close to the goal with 18.76 mm in approximately 400 steps. The results were drawn on a map (fig. \ref{fig:agent_goals}) with the goals positions and the best agent position in each episode scattered on the map (failed episodes are represented by \textbf{\textcolor{red}{red dots}} and successful episodes by \textbf{\textcolor{green}{green dots}} and each pair of goal and agent best position is connected with a blue faded line). The map shows the ability of the robot model to reach the goal effectively.
		
		\begin{table}[!htbp]
			\centering
			\caption{Performance metrics on \textbf{PPO} robot model}
			\begin{tabular}{|c|c|}
				\hline
				\textbf{Metric} & \textbf{Value} \\
				\hline
				Success Rate & 0.82 \\
				\hline
				Average Steps & 388.96 \\
				\hline
				Average Distance & 18.762 \\
				\hline
			\end{tabular}
			\label{tab:performance_metrics_robot}
		\end{table}
		
\begin{figure}[!b]
			\centering
			\includegraphics[width=0.9\linewidth]{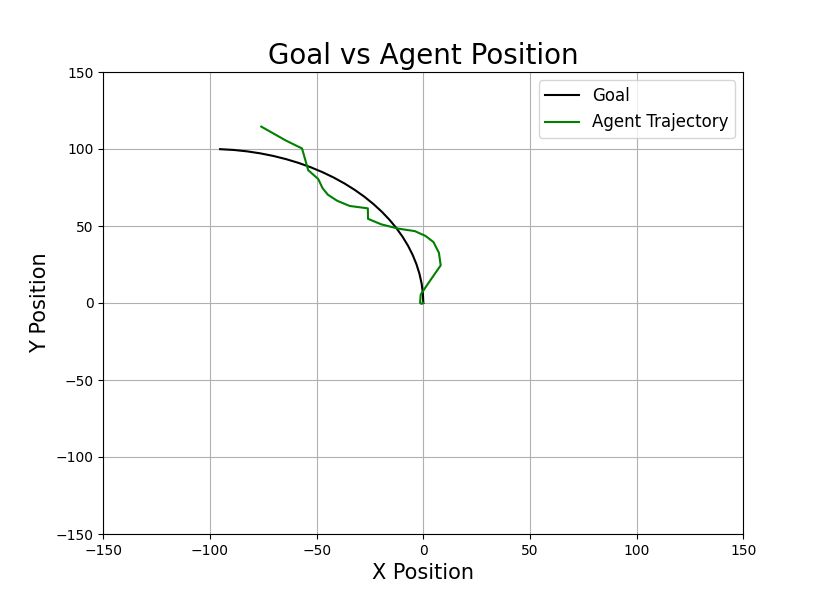}
			\caption{Map of agent following trajectory}
			\label{fig:agent_traj}
		\end{figure}

		Another test was conducted for the robot to follow a specific trajectory in the shape of an arc nearly a quarter of a circle fig. \ref{fig:agent_traj}. The robot could follow the trajectory successfully, but due to the stochastic and unpredictable behaviour of the soft materials, the robot may deviate from the original trajectory. Nevertheless, the robot stayed within the distance threshold allowed $20mm$ and reached nearly $19mm$ deviation from the path in 694 time steps.
		
		\section{Conclusion}
		\label{Conclusion}
		Once more, there is no precedent work proving its ability to model a whole soft robot inside the SOFA environment, which is very essential for the simulation stage. The main focus of this work is constructing a robust soft-legged robot environment on SOFA to let it thrive and adapt using any RL technique. This was done as explained in the previous sections by spawning our tripedal, soft-legged robot inside a well-adjusted SOFA environment and teaching it how to navigate using PPO-RL. The results were very promising, starting with the single goal reach, which reached an 82\% success rate, followed by the sequence of goals that kept on the path provided for the robot in with $19mm$ deviation from the path in just 694 time steps.
		
		For future enhancements, deploying the final learning model is important to validate its credibility in real-life scenarios similar to that in the simulation with the edit of robot state to match the sensors available. Additionally, for more improvement in the learning process, curriculum learning will be a good addition to make the agent learn step by step to achieve more complex tasks. Further development will be in the area of commercializing this robot to replace people in dangerous surroundings. This work is a kickstart chance for an uncounted number of future contributions by introducing different types of soft robots and teaching them.
		
		\addtolength{\textheight}{-12cm}   
		


		\renewcommand{\baselinestretch}{1.25} 
		
		\bibliographystyle{IEEEtran} 
		\bibliography{references} 

	\end{document}